\documentclass[11pt]{article}

\usepackage{latex/acl}

\usepackage{soul}

\usepackage{times}
\usepackage{latexsym}

\usepackage[T1]{fontenc}

\usepackage[utf8]{inputenc}

\usepackage{microtype}

\usepackage{inconsolata}

\usepackage{graphicx}

\usepackage[linesnumbered,ruled,vlined]{algorithm2e}
\usepackage{algpseudocode}
\usepackage{booktabs}    
\usepackage{amsmath}     
\usepackage{amssymb} 

\usepackage{tcolorbox}     
\usepackage{subcaption}    
\usepackage{graphicx}      
\usepackage{xcolor}   
\usepackage{multirow}
\usepackage{makecell}
\usepackage[inline]{enumitem}

\newcommand{\header}[1]{\vspace*{0.2mm}\noindent\textbf{#1}.}

\newcommand{\roxana}[1]{\textcolor{blue}{#1}}

\usepackage{latex/acl}

\SetAlFnt{\footnotesize}        
\SetAlCapFnt{\footnotesize}     
\SetAlCapNameFnt{\footnotesize} 
\SetAlgoNlRelativeSize{0}  

\title{Query Decomposition for RAG: Balancing Exploration-Exploitation}

\author{
\textbf{Roxana Petcu\textsuperscript{1}},
\textbf{Kenton Murray\textsuperscript{2}},
\textbf{Daniel Khashabi\textsuperscript{2}},
\textbf{Evangelos Kanoulas\textsuperscript{1}},
\\
\textbf{Maarten de Rijke\textsuperscript{1}},
\textbf{Dawn Lawrie\textsuperscript{2}},
\textbf{Kevin Duh\textsuperscript{2}}
\\
\\
\textsuperscript{1}University of Amsterdam,
\textsuperscript{2}Johns Hopkins University
\\
\href{mailto:r.m.petcu@uva.nl}{r.m.petcu@uva.nl}, 
\href{mailto:kenton@jhu.edu}{kenton@jhu.edu}, 
\href{mailto:danielk@cs.jhu.edu}{danielk@cs.jhu.edu}, 
\href{mailto:e.kanoulas@uva.nl}{e.kanoulas@uva.nl},
\\ 
\href{mailto:m.derijke@uva.nl}{m.derijke@uva.nl}, 
\href{mailto:lawrie@jhu.edu}{lawrie@jhu.edu}, 
\href{mailto:kevinduh@cs.jhu.edu}{kevinduh@cs.jhu.edu}
}

\newcommand{\daniel}[1]{{\color{cyan}[DK: #1]}}

\begin{document}
\maketitle
\begin{abstract}
Retrieval-augmented generation (RAG) systems address complex user requests by decomposing them into subqueries, retrieving potentially relevant documents for each, and then aggregating them to generate an answer.
Efficiently selecting informative documents requires balancing a key trade-off: (i) retrieving broadly enough to capture all the relevant material, and (ii) limiting retrieval to avoid excessive noise and computational cost.
We formulate query decomposition and document retrieval in an exploitation-exploration setting, where retrieving one document at a time builds a belief about the utility of a given sub-query and informs the decision to continue exploiting or exploring an alternative. We experiment with a variety of bandit learning methods and demonstrate their effectiveness in dynamically selecting the most informative sub-queries.
Our main finding is that estimating document relevance using rank information and human judgments yields a 35\% gain in document-level precision, 15\% increase in \(\alpha\text{-nDCG}\), and better performance on the downstream task of long-form generation. Code is available on GitHub.\footnote{\url{https://anonymous.4open.science/r/query-decomposition-bandits-2A0D}}
\end{abstract}

\section{Introduction}
\label{sec:introduction}

\begin{figure*}[t]
    \includegraphics[width=\linewidth]{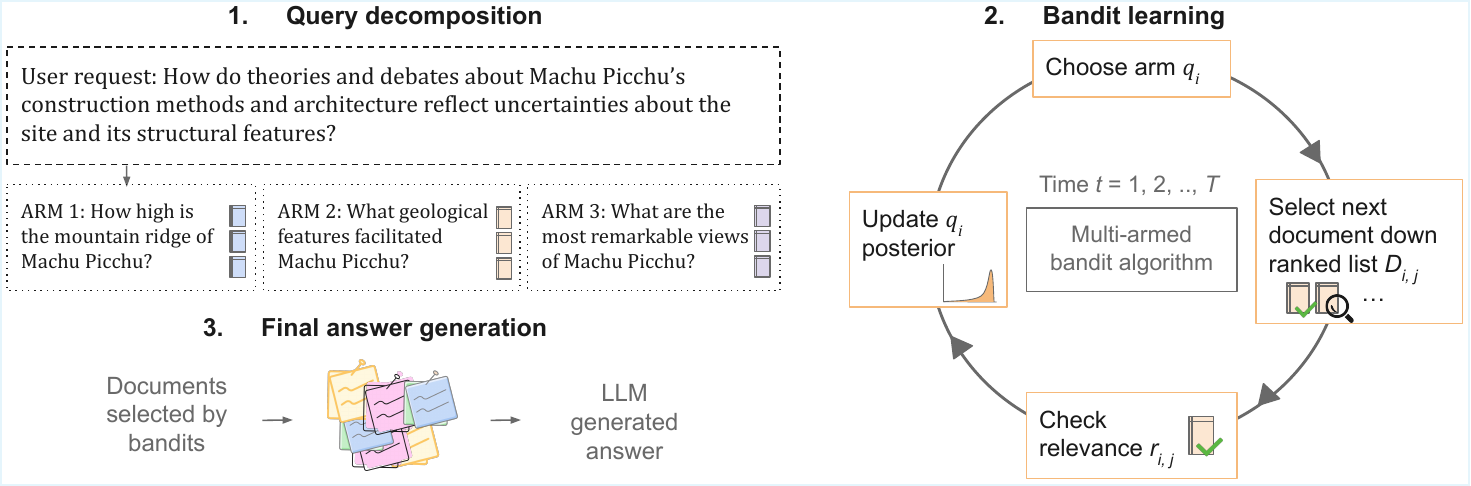}
\caption{A user request is first decomposed into sub-queries. Bandit learning iteratively selects a sub-query (arm), retrieves a document down its ranked list, observes its relevance, and updates the sub-query posterior belief over time. The selected documents across iterations are then used as evidence in a RAG setting to generate a grounded answer, balancing exploration (more sub-queries) and exploitation (more documents per sub-query).}
    \label{fig:query_decomposition_task}
\end{figure*}

Complex user queries usually involve discourse operators such as the exclusion of information \cite{excluir}, negation \cite{petcu2025comprehensivetaxonomynegationnlp, nevir, repro_nevir}, or missing entities \cite{reasoning_missing_entities, tip_of_the_tongue}, and often require retrieving evidence found in multiple documents.
One way to handle them is to decompose the request into atomic sub-queries~\cite{khot2023decomposedpromptingmodularapproach}, as shown in Figure~\ref{fig:query_decomposition_task}. 
Retrieving documents independently for each sub-query leads to coverage of complex information needs by maximizing recall. However, it may result in a large number of documents, of which many are irrelevant \cite{kim2025NotInformative}. This is problematic in two ways. 
First, too many documents cannot fit in the context window of an LLM, without being arbitrarily truncated, potentially removing vital information.
Second, including a large proportion of irrelevant documents introduces noise in the generation \cite{jin2024longcontextllmsmeetrag}. 
Mitigating these problems involves filtering by either human annotators or LLM agents. This motivates our core question: \emph{how to efficiently identify which sub-queries are likely to retrieve relevant information and what constitutes an appropriate retrieval depth for each sub-query}?

Existing approaches to sub-query decomposition and document retrieval lack a principled mechanism for allocating documents, with most methods retrieving a fixed number of documents regardless of sub-query utility. Meanwhile, we want to study a method that adaptively decides, under a fixed budget, whether to continue retrieving documents from an observed, promising sub-query or to explore alternatives that may yield more relevant evidence, while avoiding irrelevant overlap. A multi-armed bandit framework naturally captures this process: each sub-query is treated as an arm, for which each observed document provides evidence of its utility.
Framing query decomposition and document selection as a bandit problem  addresses two core challenges of complex information needs. First, retrieval is inherently sequential and budget-constrained, as it is impossible to verify all documents. Second, the relevance of a sub-query is initially uncertain, while our belief of its relevance builds with each retrieved document.

Figure \ref{fig:query_decomposition_task} illustrates our setting. We estimate the utility of each sub-query, modeled as an arm, by observing the relevance of one document at a time. The choice of assessment directly influences the cost. By estimating utility under a fixed budget, i.e., at each step the system makes a choice between going down the ranked list of a certain sub-query or retrieving from a new one, the process becomes considerably more efficient. This perspective is complementary to recent approaches on efficient selection under computational constraints, such as compute-efficient re-ranking \citep{SetwiseInsertion2024} and training data selection \citep{petcu2024efficientdataselectionemploying}. In the example shown in Figure \ref{fig:query_decomposition_task} 
we assume access to a user request, its decomposition, document relevance (assessed by either LLM judges or human annotators), and a search engine 
We aim to answer the following research questions:

\begin{enumerate}[label=(RQ\arabic*),leftmargin=*,nosep]
    \item How can query selection be framed as a multi-armed bandit problem? Does bandit learning outperform full exploitation and full exploration strategies? What strategies best balance exploration with exploitation?
    \item Does document selection using multi-armed bandits improve evaluation metrics on the downstream task of report generation?
    \item Can bandit learning be used to guide hierarchical sub-query decomposition?
\end{enumerate}

\noindent%
For answering RQ1, we model query decomposition and document retrieval in an efficient way using reinforcement learning (RL) policies in a multi-armed bandit setting, demonstrating that estimating Bernoulli distributions boosts relevance estimates by \(17\%\) compared to simply going down the ranked list. With RQ2, we look into the performance of using an optimal subset of documents for report generation, in which we improve on evaluation metrics such as nugget coverage and sentence support. With RQ3, we examine the use of hierarchical, multi-level sub-query decomposition, which yields a \(30\%\) precision gain over selecting all documents from a single-level decomposition.
\section{Related Work}

\textbf{RAG systems.}
RAG systems extend language models with access to external knowledge from retrieved evidence \cite{lewis2021retrievalaugmentedgenerationknowledgeintensivenlp, asai2020multipledocuments, soudani2024surveyrecentadvancesconversational}. While this approach brings advantages in grounding the generation into real information \cite{askari2025selfseedingmultiintentselfinstructingllms}, it treats the query as a unitary piece of information. There are variants of RAG systems that handle complex queries, such as multi-step RAG systems which perform iterative retrieval \cite{Gu2025ChainOfRetrieval}, or feedback-based retrieval implemented with RLHF or model-estimated policies for re-generating the query until a satisfying answer can be composed by the system \cite{deng2022rlpromptoptimizingdiscretetext, rafailov2024DPO, Jun2025SearchR1}. This line of work follows earlier efforts in RL-based query reformulation \cite{Buck2018ReformulationQuery}, which have been extended to multi-modal and retrieval tasks \cite{odijkMaarten}.

\header{Query decomposition} 
Research in discourse phenomena such as exclusion \cite{excluir}, negation \cite{nevir, petcu2025comprehensivetaxonomynegationnlp}, and compositions of logic operators \cite{Zhang2024BoolQuestionsDD} shows that such formulations are difficult to encode by retrieval models \cite{krasakis2025constructingsetcompositionalnegatedrepresentations}. These cases often benefit from decomposing the original request into simpler, unitary sub-queries \cite{yaoreasoningpaths}. Query decomposition can be performed semantically using external models \cite{khot2023decomposedpromptingmodularapproach}.
However, naively splitting a complex query and retrieving documents for each sub-query does not necessarily improve results: longer contexts can degrade downstream performance \cite{shao2025reasonirtrainingretrieversreasoning}. To address this, we propose an adaptive allocation of retrieval budgets across sub-queries. 

\header{Research gap} We position our work at the intersection of query decomposition, using LLMs for unitary information needs, and efficient retrieval of documents by observing and modeling relevance-aware distributions for each decomposed sub-query. The closest existing work applies multi-armed bandits and active learning techniques for document selection for large-scale evaluation, and for building fair IR test collections~\cite{Li_2017, Rahman_2020, 10.1145/3269206.3271766}.
\section{Methodology}

\subsection{The exploitation-exploration problem}



The exploration-exploitation dilemma is a fundamental decision-making concept that balances the act of exploiting known low-risk options with exploring unknown high-risk alternatives. A popular paradigm for this setting is the multi-armed bandit problem, which assumes access to multiple fixed choices, called arms, observed iteratively by a decision maker. The properties of each arm are initially uncertain, and the belief about their relevance is refined as more evidence is observed. A fundamental aspect of the bandit problem is that sampling from an arm does not affect the underlying distribution of that arm or any other. It can be seen as a set of real distributions \(B = \{R_1, R_2, \ldots, R_K\}\), where each observed value is associated with a reward.


\begin{figure}[!t]
    \centering
    \includegraphics[width=\linewidth]{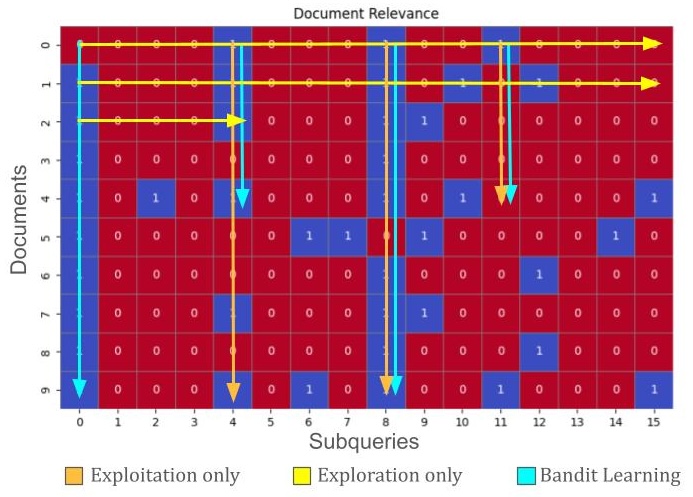}
    \caption{Exploration–exploitation across \(n\) subqueries and \(m\) documents; colors indicate document relevance, and arrows show how exploitation, exploration, and bandit-based policies allocate effort across sub-queries.}
    \label{fig:informative_subqueries}
\end{figure}

\subsection{Bandit learning for query decomposition}
Figure \ref{fig:informative_subqueries} illustrates the setting of our problem: given a user request \(\mathcal{Q}\) and its decomposition \(\{q_1, q_2, \ldots, q_K\}\), we retrieve a ranked list of \(N\) documents for each sub-query \(q_i\) as \(\{d_{i, 1}, d_{i, 2}, \ldots, d_{i, N}\}\). We observe one document at a time -we retrieve a document, assess its relevance with respect to \(\mathcal{Q}\), and store it as an evaluated document- while trying to maximize observed relevance while not going over a budget of \(b\) documents, where \(b < N\cdot K\). We model each sub-query \(q_i\) as an arm, where its associated set \(\mathcal{D}_i\) has an unknown distribution over relevance labels w.r.t.\ \(\mathcal{Q}\). We treat each arm as an unknown distribution over \(\mathcal{D}_i\) for which we maintain a posterior over its expected utility. We initialize the distribution with an uninformative (flat) prior and update the posterior after each observation over \(\mathcal{D}_i\). We employ Thompson sampling, a standard multi-armed bandits (MAB) algorithm, both in discrete (Algorithm \ref{tab:thompson_discrete_space}) and continuous space (Algorithm \ref{tab:thompson_continuous_space}).





    



\begin{algorithm}[!t]
\footnotesize
\caption{Thompson Sampling in Discrete Space}
\label{tab:thompson_discrete_space}

\SetKwInOut{Input}{Input}
\SetKwInOut{Output}{Output}
\Input{Sub-queries $\mathcal{S}=\{s_1,\dots,s_K\}$; Budget $b$}
\BlankLine

\For{$i=1,\dots,K$}{
  Define Beta priors $\alpha_i \gets 1,\; \beta_i \gets 1$
}

Initialize observation set $\mathcal{O} \gets \emptyset$\;

\For{$t=1,\dots,b$}{
  \For{$i=1,\dots,K$}{
    Sample $\theta_i \sim \mathrm{Beta}(\alpha_i,\beta_i)$
  }
  Select arm $a_t \gets \arg\max_i \theta_i$\;

  Pick next document $d_{a_t,n} \in \mathcal{D}_{a_t}$ where\\
  \hspace{1em}$n \gets \min\{\, j\in\{1,\ldots,N\}\mid (a_t,j)\notin\mathcal{O}\,\}$\;

  Observe reward $r(a_t,n)$\;

  Update posterior:\\
  \hspace{1em}$\alpha_{a_t} \gets \alpha_{a_t} + r(a_t,n)$; \quad
  $\beta_{a_t} \gets \beta_{a_t} + (1 - r(a_t,n))$\;

  Save observation $\mathcal{O} \gets \mathcal{O} \cup \{(a_t,n)\}$\;
}
\end{algorithm}

In discrete space, we model a Bernoulli distribution with a Beta conjugate prior, while in continuous space, we model Gaussians with Gaussian priors. Importantly, (i) we model the multi-armed bandit problem on query decomposition, where the arm chosen at time \(t\) is denoted as \(a_t\), (ii) each observation is represented by the next document \(d_{:,j}\) down a ranked list,
and (iii) the reward is calculated at the document level, i.e., \(d_{{a_t}, j}\) for the chosen arm \(a_t\) and document rank \(j\). 

\subsection{Methods and rewards}\label{sec:rewards}
We study different properties that we hypothesize to play an important role in the task of sub-query-dependent document retrieval for RAG:

\noindent\textbf{Rank information}: Each sub-query \(q_i\) is associated with a ranked list of retrieved documents, whose order encodes an implicit estimate of relevance. To incorporate this rank-based signal, we calculate the reward for the sub-query \(a_t\) chosen at time \(t\) as the average relevance over a local window of \(k\) documents down the ranked list: \(\frac{1}{k} \sum_{i=n}^{n+k-1} \text{Relevance}(d_{a_t, i})\), where \(d_{a_t, i}\) represents the \(i\)-th document retrieved for subquery chosen at time \(t\), \(n\) is the current position in the ranked list, and $\text{Relevance}(d_{a_t, i})$ represents the relevance of one document, either as a binary label (from human or LLM judges), or a continuous relevance signal (ranking score).

\noindent\textbf{Diversity of documents}: We want the retrieved documents to be diverse; retrieving a relevant document that has a high content overlap with a previously observed one does not bring new information. To encourage this, we estimate the novelty of the currently observed document \(d_{a_t, i}\) with respect to the set of previously observed documents \(\mathcal{O}\); we model this by estimating the novelty of each sub-query using cosine similarity: \(1 - \frac{\max_{(i, j) \in \mathcal{O}} \cos(d_{a_t, n}, d_{i, j}) + 1}{2}\).

\noindent\textbf{Exploration}: We want to explore each sub-at least once; we enforce this using an upper confidence bound (UCB) term \(c \cdot \sqrt{\frac{1}{n}\log_2(n + 1)}\), with \( c \to 0 \). This term diverges to infinity for subqueries with no observations, i.e., when \(n\rightarrow0\).

\noindent%
These factors compose our final reward policy, which we propose as a Bernoulli top-$k$ UCB diversity-aware estimate in Eq.~\ref{eq:top_reward}, used on line 9 in Algorithm \ref{tab:thompson_discrete_space}: 
\begin{equation}
\begin{split}
r(a_t, & \,n) = {}\\
& \frac{1}{k} \sum_{i=n}^{n+k-1} \text{Relevance}(d_{a_t, i}) \cdot {} \\
& \left( 1 -
    \frac{\max\limits_{(i, j) \in \mathcal{O}} \cos(d_{a_t, n}, d_{i, j}) + 1}{2} \right) \\
& {}+ c \cdot \sqrt{\frac{\log_2(n + 1)}{n}}, 
 \text{ with } c \to 0.
\end{split} 
\label{eq:top_reward}
\end{equation}

\subsection{Hierarchical query decomposition with Correlated MAB}\label{sec:hierarchical_method}

\begin{figure}[!t]
    \includegraphics[width=\linewidth]{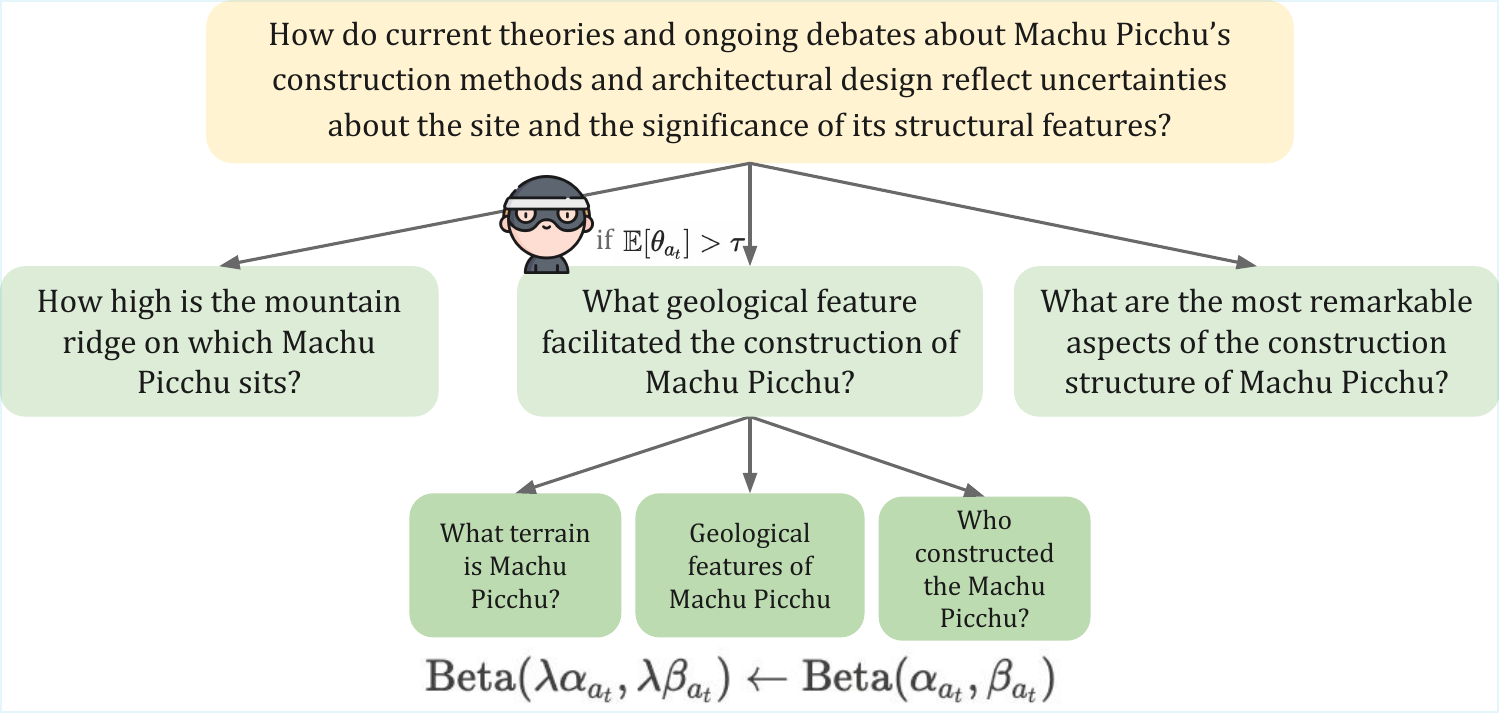}
    \caption{Hierarchical query decomposition with correlated bandits, where informative sub-queries are decomposed into arms with inherited posterior beliefs.}
    \label{fig:query_decomposition_herarchical}
\end{figure}

In hierarchical decompositions, a complex user request is iteratively split into smaller subqueries, where each can be further decomposed into more fine-grained information needs (see Figure \ref{fig:query_decomposition_herarchical}). As a result, the retrieval space forms a hierarchy of sub-queries and their associated document distributions, where each child sub-query inherits properties from its parent. As some sub-queries may be more informative than others, instead of uniform expansion, we selectively expand those that demonstrate high utility according to their posterior estimates. A sub-query \(q_i\) is highly informative when its estimated value is above a set informativeness threshold \(\mathbb{E}_{q_i} = \frac{\alpha_i}{\alpha_i + \beta_i} > \tau\) and when it has been observed at least \(n\) times. If an informative sub-query \(q_i\) is expanded, its children \(\{q_{i, 1}, q_{i, 2}, ..., q_{i, m}\}\) become new arms in the multi-armed bandit formalization, whose distributions are correlated with their parent sub-query \(q_i\). This correlation is captured by an inheritance factor \(\lambda \in (0, 1]\), i.e. a parent with posterior \(\text{Beta}(\alpha_i, \beta_i)\) will yield a child sub-query initialized as \(\text{Beta}(\lambda \alpha_i, \lambda \beta_i)\). 

\section{Experimental Setup}

\begin{table*}[!t]
\centering
\setlength{\tabcolsep}{1.5pt}
\resizebox{\textwidth}{!}{%
\begin{tabular}{ll}
    \toprule
    \textbf{Reward type} & \textbf{Formula} \\
    
    \midrule

    Baseline random & \(r(a_t) = \text{Relevance}(d_{a_t, n})\) \text{ where} \(a_t \sim \text{unif}(a_1, ..., a_K), n \sim \text{unif}(rank_1, rank_2, .. rank_m)\) \\[0.5em]
    
    Baseline random rank-aware & \(r(a_t, n) = \text{Relevance}(d_{a_t}, n)\) \text{ where} \(a_t \sim \text{unif}(a_1, ..., a_K)\) \\[0.5em]

    \midrule
    \(\epsilon\)-greedy & \(r(a_t, n) = \text{Relevance}(d_{a_t, n})\) \text{ if} \(\text{Relevance}(d_{a_t, n-1})\) \text{ is 1 else } \(a_t \sim \text{unif}(a_1, ..., a_K)\) \\[0.5em]
    
    Bernoulli & \( r(a_t, n) = \text{Relevance}(d_{a_t, n}) \) \\[0.5em]
    
    Bernoulli UCB & \( r(a_t, n) = \text{Relevance}(d_{a_t, n}) + c \cdot \sqrt{\frac{\log_2(n + 1)}{n}} \quad \text{with } c \to 0 \) \\[0.5em]
    
    Bernoulli top-\(k\) & \( r(a_t, n) = \frac{1}{k} \sum_{i=n}^{n+k-1} \text{Relevance}(d_{a_t, i}) \) \\[0.5em]
    
    Bernoulli rank-aware & \( r(a_t, n) = \frac{\text{Relevance}(d_{a_t, n})}{\log_2(n + 2)} \) \\[0.5em]
    
    Gaussian & \( r(a_t, n) = \text{RFF}(d_{a_t, n}) \quad \text{or} \quad \text{ColBERT}(d_{a_t, n}) \) \\[0.5em]
    
    Diversity & \( r(a_t, n) = \text{Relevance}(d_{a_t, n}) \cdot \left(1 - \frac{\max\limits_{(i, j) \in \mathcal{O}} \cos(d_{a_t, n}, d_{i, j}) + 1}{2}\right) \) \\[0.5em]
    
    Diversity concave & \(
    r(a_t, n) = \text{Relevance}(d_{a_t, n}) \cdot 
    \begin{cases}
    1, & \text{if } \max\limits_{j < n} \cos(d_{a_t, n}, d_{a_t, j}) < 0 \\
    \exp\left(-a \cdot \left(\max\limits_{(i, j) \in \mathcal{O}} \cos(d_{a_t, n}, d_{i, j}))\right)^b\right), & \text{otherwise}
    \end{cases}
    \)  \\[0.5em]

    Bernoulli top-\(k\) UCB diversity & \( r(a_t, n) = \frac{1}{k} \sum_{i=n}^{n+k-1} \text{Relevance}(d_{a_t, i}) \cdot \left(1 - \frac{\max\limits_{(i, j) \in \mathcal{O}} \cos(d_{a_t, n}, d_{i, j})) + 1}{2}\right) + c \cdot \sqrt{\frac{\log_2(n + 1)}{n}}, \text{ with } c \to 0 \)\\
    \bottomrule
    \end{tabular} }
    \caption{Reward policies used for evaluating sub-query selection strategies. For all but the baseline random, we assume \(n = \min \{ j \in \{1,\ldots,N\} \mid (a_t, j) \notin \mathcal{O} \}\). For the diversity concave policy, we assume \(a=5\) and \(b=15\).}
\label{tab:reward_functions}
\end{table*}

We run our experiments on two datasets, which come with different properties and challenges. 

\subsection{NeuCLIR}
\header{Dataset}
The \textbf{NeuCLIR} dataset \cite{lawrie2025overviewtrec2024neuclir} features decomposable user requests; the sub-queries are generated in a serialized manner, i.e., a one-level decomposition. NeuCLIR is part of TREC (Text Retrieval Conference) and is designed to evaluate information retrieval models on multilingual data. More precisely, the dataset contains complex user requests and a corpus of documents in Chinese ($\sim$3M documents), Russian ($\sim$5M documents), and Persian (2M documents). \footnote{\url{https://huggingface.co/neuclir}} For the purpose of this study, all documents have been translated into English. We run our experiments on the entirety of the human-annotated nugget partition of the dataset. The average length of each user request is \(51.95 \pm 19.46\) words.

\header{Retrieval} We decompose NeuCLIR user requests into \(k=16\) sub-queries using LLM calls as specified in prompts \ref{fig:problem-definition-prompt} and \ref{fig:specialized-information-prompt}. For each generated sub-query, we run a search engine to retrieve \(n=10\) documents. The retriever used in this study is a combination of PLAID-X, a dual encoder with late interaction \cite{colbert-x, translate-distill}, learned sparse retrieval (LSR) \cite{LSRandrewThong}, which combines sparse retrieval with contextualized dense embeddings, and a Qwen retriever \cite{bai2023qwentechnicalreport, qwen2025qwen25technicalreport}.

\begin{figure*}[t!]
    \centering
    \includegraphics[width=\linewidth]{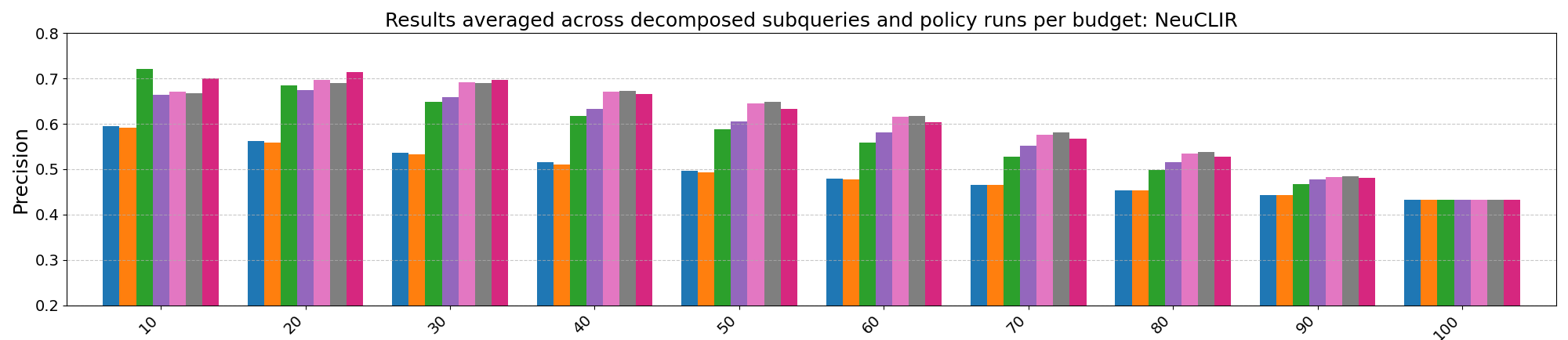}
    \vspace{1em}  
    \includegraphics[width=\linewidth]{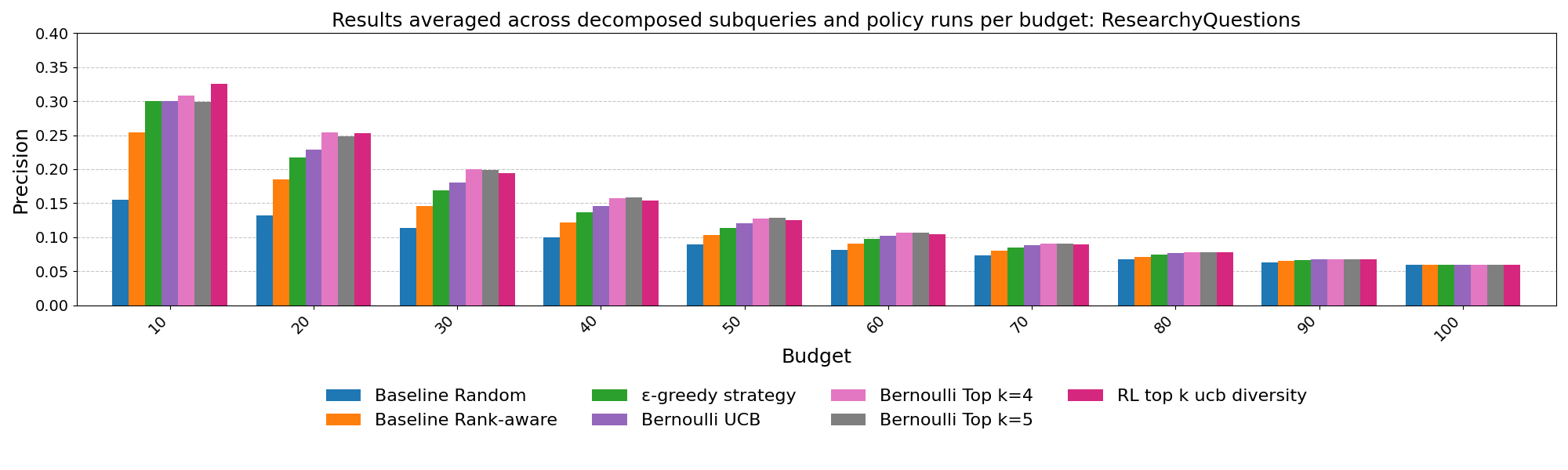}
    \vspace*{-13mm}
    \caption{Precision scores on NeuCLIR24 and ResearchyQuestions for baselines and reward policies, evaluated across varying budgets of observed documents. The budgets are expressed as percentages of the total available corpus. Modelling the queries using Bernoulli distribution considering the top ranked k documents performs the best across budgets, with all policies reaching the same performance as the budget covers the whole space of actions.}
    \label{fig:neuclir_precision}
\end{figure*}

\subsection{ResearchyQuestions}
\textbf{Dataset.} 
ResearchyQuestions \cite{rosset2024researchyquestionsdatasetmultiperspective} is composed of \(100k\) complex Bing questions that are non-factoid, multi-perspective, and require decomposition. Each instance in the dataset is decomposed into sub-queries hierarchically over two levels: 
\textit{headers} (first-level sub-queries), and \textit{sub-queries} (second-level). We build the corpus by aggregating all documents marked as relevant across the entire ResearchyQuestions dataset. 

\header{Retrieval} We apply BM25 to retrieve top 10 documents for each query. Since some instances have limited document coverage, we filter the data and keep only those where at least 10\% of the retrieved documents are labeled as relevant. The final dataset used for evaluation consists of 140 instances.


\subsection{Implementation Details}

\header{Rewards} We experiment with the reward introduced in Equation \ref{eq:top_reward} and with several alternatives and baselines described in Table \ref{tab:reward_functions}. We sample documents up to a budget \( b \in \{10\%, 20\%, \ldots, 100\%\} \) out of the total document set of size \(N \times K\). For \(b = 100\%\) we expect all policies and baselines to converge to the same performance. To eliminate noise from random policy starts, we run each experiment \(1000\) times. Moreover, as we decompose the user request using LLM calls, we run the decomposition \(10\) times for each user request, leading to a total of $19 \times 10 \times 1000$ experiments over which we average for each budget. For the Gaussian reward, we experiment with both the RFF and PLAID-X ranking scores, depending on the search engine used. For the diversity concave reward, we set hyperparameters \(a = 5\) and \(b = 15\). 

\header{Metrics} We evaluate performance using precision and \(\alpha\)-nDCG \cite{alphandcg} for document selection. 
For long-form report generation, we use the Auto-ARGUE framework \cite{autoargue}.

\section{Empirical Results}

Our experiments are designed to answer the research questions presented in Section \ref{sec:introduction}. We answer RQ1 (can we frame document selection as a multi-armed bandit problem?) through experiments on NeuCLIR and ResearchyQuestions in Section \ref{section:neuclir_eval}. We answer RQ2 (do policy-observed documents lead to better performance on downstream tasks?) by long-form generation of reports on NeuCLIR in Section \ref{section:neuclir_downstream_task}. We answer RQ3 (can we apply bandit-learning on hierarchical sub-queries?) by modeling correlated bandits on existing sub-query splits from ResearchyQuestions in Section \ref{section:researchy_correlated}.

\subsection{Document selection as a multi-armed bandit problem}\label{section:neuclir_eval}

\begin{table*}[!ht]
  \centering
  \setlength{\tabcolsep}{5pt}      
  \renewcommand{\arraystretch}{1.2} 
  \resizebox{\linewidth}{!}{%
  \begin{tabular}{l*{6}{c}}
    \toprule
    & \multicolumn{3}{c}{Budget $b=20\%$} & \multicolumn{3}{c}{Budget $b=30\%$} \\
    \cmidrule(lr){2-4} \cmidrule(lr){5-7}
    & \makecell{Citation\\Support} 
    & \makecell{Nugget\\Coverage} 
    & \makecell{Sentence\\Support} 
    & \makecell{Citation\\Support} 
    & \makecell{Nugget\\Coverage} 
    & \makecell{Sentence\\Support} \\
    \midrule
    Full   & 0.788 & 0.461 & 0.780 & 0.788 & 0.461 & 0.780 \\
    \midrule
    Random               & $0.826 \pm 0.043$ & $0.477 \pm 0.055$ & $0.822 \pm 0.045$ & $0.827 \pm 0.054$ & $0.446 \pm 0.052$ & $\textbf{0.837} \pm \textbf{0.055}$ \\
    Rank                 & $0.834 \pm 0.038$ & $0.462 \pm 0.045$ & $0.819 \pm 0.045$ & $0.818 \pm 0.053$ & $0.462 \pm 0.060$ & $0.791 \pm 0.067$ \\
    \midrule
    $\epsilon$-greedy    & $0.823 \pm 0.042$ & $0.490 \pm 0.053$ & $0.809 \pm 0.049$ & $0.799 \pm 0.071$ & $0.478 \pm 0.080$ & $0.775 \pm 0.077$ \\
    Bernoulli UCB        & $0.834 \pm 0.031$ & $0.477 \pm 0.042$ & $0.823 \pm 0.037$ & $0.830 \pm 0.051$ & $0.497 \pm 0.071$ & $0.798 \pm 0.080$ \\
    Bernoulli $k=4$      & $\textbf{0.866} \pm \textbf{0.037}$ & $0.463 \pm 0.044$ & $\textbf{0.855} \pm \textbf{0.039}$ & $0.834 \pm 0.060$ & $0.480 \pm 0.075$ & $0.817 \pm 0.075$ \\
    Bernoulli $k=5$      & $0.849 \pm 0.035$ & $\textbf{0.492} \pm \textbf{0.045}$ & $0.839 \pm 0.038$ & $0.831 \pm 0.056$ & $\textbf{0.500} \pm \textbf{0.064}$ & $0.819 \pm 0.072$ \\
    Top-K UCB Div.  & $0.837 \pm 0.031$ & $0.463 \pm 0.040$ & $0.836 \pm 0.030$ & $\textbf{0.835} \pm \textbf{0.046}$ & $0.475 \pm 0.050$ & $0.826 \pm 0.061$ \\
    \bottomrule
  \end{tabular}
  }
  \caption{Report Generation results for budgets \(b=20\%\) and \(b=30\%\) with their confidence intervals. Selecting a relevant subset of documents boosts downstream report generation performance with \(6.0\)--\(9.9\%\) increase in citation support, \(6.7\)--\(8.5\%\) in nugget coverage, and \(7.3\)--\(9.6\%\) in sentence support.}
  \label{tab:neuclir_report_generation_b20_b30}
\end{table*}

Figure \ref{fig:neuclir_precision} illustrates macro-precision over document budget. We highlight the main findings in Figure \ref{fig:neuclir_precision}, while Figures \ref{fig:neuclir_precision_all} and \ref{fig:researchy_eval_all} in the Appendix present all results. Exploitation and exploration-only policies perform similarly to our baselines: the exploitation-only policy returns a precision of \(0.57\) and exploration-only a precision of \(0.55\) on NeuCLIR, while both exploitation and exploitation-only achieve \(0.14\) precision on ResearchyQuestions. Based on results on our proposed rewards, we observe that: (i) reward policies effectively model sub-query relevance; (ii) \(\epsilon\text{-greedy}\) performs best when using \(10\)--\(20\%\) of the budget, likely because early exploitation of high-relevance arms is optimal; and (iii) using rank-information gives a noticeable advantage. All rewards achieve the same macro-precision as the budget reaches maximum value, due to coverage of the same set of documents by all policies and baselines. Figure \ref{fig:neuclir_recall_all} (in the Appendix) shows similar insights on recall w.r.t.\ the total amount of relevant documents retrieved for all sub-queries.

\begin{table}[ht]
  \centering
  \setlength{\tabcolsep}{3.5pt}
  \renewcommand{\arraystretch}{1.2}
  \resizebox{\linewidth}{!}{%
  \begin{tabular}{l*{5}{c}*{2}{c}}
    \toprule
     & \multicolumn{5}{c}{\makecell{\textbf{\(\alpha\text{-nDCG}\)} $\uparrow$ \\ NeuCLIR24}} 
     & \multicolumn{2}{c}{\makecell{\textbf{\(\alpha\text{-nDCG}\)} $\uparrow$ \\ Researchy}} \\
    \cmidrule(lr){2-6} \cmidrule(lr){7-8}
     & $K{=}5$ & $K{=}10$ & $K{=}20$ & $K{=}40$ & $K{=}50$ & $K{=}5$ & $K{=}10$ \\
    \midrule
    Random               & 0.411 & 0.454 & 0.479 & 0.263 & 0.311 & 0.134 & 0.127 \\
    Rank                 & 0.426 & 0.462 & 0.476 & 0.265 & \underline{0.313} & 0.172 & 0.137 \\
    \midrule
    $\epsilon$-greedy    & \textbf{0.500} & \textbf{0.539} & 0.546 & \textbf{0.286} & \textbf{0.314} & 0.207 & 0.145 \\
    Bernoulli            & 0.445 & 0.508 & 0.543 & 0.246 & 0.282 & 0.210 & \underline{0.157} \\
    Bernoulli UCB        & 0.460 & 0.517 & 0.545 & 0.248 & 0.279 & 0.210 & \textbf{0.164} \\
    Bernoulli $k=4$      & 0.461 & 0.536 & 0.551 & \underline{0.275} & 0.283 & \textbf{0.240} & 0.153 \\
    Bernoulli $k=5$      & 0.448 & 0.523 & \textbf{0.555} & 0.267 & 0.291 & \underline{0.232} & 0.156 \\
    Bernoulli Rank       & 0.437 & 0.481 & 0.523 & 0.264 & 0.289 & 0.208 & 0.156 \\
    Diversity            & 0.444 & 0.505 & 0.541 & 0.260 & 0.281 & 0.210 & 0.154 \\
    Diversity Conc.      & 0.442 & 0.512 & 0.540 & 0.260 & 0.283 & 0.211 & 0.149 \\
    Top-$k$ UCB Div.  & \underline{0.469} & \underline{0.537} & \underline{0.552} & 0.253 & 0.285 & 0.222 & 0.154 \\
    \bottomrule
  \end{tabular}
  }
  \caption{Comparison of $\alpha$-nDCG over different budgets \(k\) for NeuCLIR24 and ResearcyQuestions. Bernoulli posteriors are the best to estimate document relevance regardless the value of \(k\) for $\alpha$-nDCG, with an additional performance boost when considering the top k ranked documents. UCB also brings an advantage. Div. in the last metrics refers to Diversity.}
  \label{tab:neuclir_alphandcg}
\end{table}


Table \ref{tab:neuclir_alphandcg} shows results for multiple values of \(\alpha\text{-nDCG}\). We observe that \(\alpha\text{-nDCG}\) achieves higher values for baselines and all policies for budget \(K \in [5, 10, 20]\). As the budget grows, all baselines and policies converge towards the same pool of documents, reducing marginal novelty, as \(\alpha\text{-nDCG}\) punishes redundant content. Out of all policies, notably the \(\epsilon\text{-greedy}\), UCB, and ranked Bernoulli rewards perform the best. 

We frame query decomposition and document selection as a stochastic multi-armed bandit where each subquery is an independent arm, and pulling an arm corresponds to observing a document down the ranked list. Each observation receives a Bernoulli reward based on relevance, updating the subquery’s relevance distribution. Across experiments, we show that bandit learning outperforms full exploitation and exploration, and that a policy combining rank-information, diversity, and an upper confidence bound (Equation \ref{eq:top_reward}) outperforms other variants in Table \ref{tab:reward_functions}.

\subsection{Impact on report generation}\label{section:neuclir_downstream_task}


We generate reports on NeuCLIR24 using document subsets retrieved by our policies for budgets \(b=[20\%;30\%]\), which provided the best trade-off between evidence coverage and precision in preliminary experiments (Figure~\ref{fig:neuclir_precision}). Evaluations are conducted with the Auto-ARGUE framework, which measures four aspects of grounded generation: citation relevance (quality of cited sources), citation support (alignment between claims and citations), nugget coverage (proportion of unique information units), and sentence support (share of sentences with appropriate evidence).
An example report is provided in Appendix~\ref{sec:app_report_gen_example}.
As shown in Table~\ref{tab:neuclir_report_generation_b20_b30}, top-$k$ Bernoulli policies achieve the strongest performance in both nugget coverage and factual support. Citation relevance is omitted, as it remains constant across policies. Results show that selecting documents using multi-armed bandits improves downstream report generation. The strongest performance gains follow the rank-aware top-$k$ Bernoulli policies, including top-$k$ UCB with diversity \emph{citation support} at $b{=}30\%$ ($0.835$).

\subsection{Bandit learning for hierarchical sub-queries}\label{section:researchy_correlated}

We use ResearchyQuestions to evaluate our correlated bandits formulation. Results in Table \ref{tab:budget10_results} show clear improvements over policies for a small budget \(b = 10\%\). 
We pick the three hyperparameters described in Section \ref{sec:hierarchical_method} through Bayesian hyperparameter search. The results are presented in Figure \ref{fig:bayesian_lookup} in the Appendix, given which we chose a minimal number of observations \(n=4\), informativeness threshold \(\tau=0.77\), and inheritance factor \(\lambda=0.91\).

Table \ref{tab:budget10_results} shows that we can use bandit-learning to model hierarchical query decomposition. For budget \(b=10\%\), there is a clear performance boost when using the hierarchical structure compared to treating all subqueries at the same serial level. 
Performance using Bernoulli increases by \(21.6\%\), while top-\(k=5\) has a performance boost of \(32.3\%\), and RL top-\(k\) UCB diversity a gain of \(20.4\%\). As the budget increases, the gains converge. These results indicate that hierarchy-aware (correlated) bandits best balance exploration/exploitation early by focusing on promising branches of the sub-query tree, yielding higher precision under small budgets.

\begin{table}[ht]
  \centering
  \setlength{\tabcolsep}{4.5pt}
  \renewcommand{\arraystretch}{1.2}
  \resizebox{\linewidth}{!}{%
  \begin{tabular}{l *{2}{c} *{2}{c}}
    \toprule
     & \multicolumn{2}{c}{\textbf{\(b = 10\%\)}} & \multicolumn{2}{c}{\textbf{\(b = 20\%\)}} \\
    \cmidrule(lr){2-3} \cmidrule(lr){4-5}
     & Ser. & Hier. & Ser. & Hier. \\
    \midrule
    Baseline Random              & 0.157 & 0.159 & 0.137 & 0.136 \\
    Baseline Rank-aware          & 0.263 & 0.264 & 0.191 & 0.193 \\
    $\epsilon$-greedy strategy   & 0.306 & 0.306 & 0.223 & 0.223 \\
    Bernoulli                    & 0.305 & \textbf{0.371} & 0.237 & 0.246 \\
    Bernoulli UCB                & 0.306 & \textbf{0.368} & 0.234 & 0.245 \\
    Bernoulli Top $k=3$            & 0.325 & \textbf{0.391} & 0.260 & 0.250 \\
    Bernoulli Top $k=4$            & 0.317 & \textbf{0.398} & 0.261 & 0.2512 \\
    Bernoulli Top $k=5$            & 0.303 & \textbf{0.401} & 0.257 & 0.250 \\
    Bernoulli Rank-Aware         & 0.304 & \textbf{0.375} & 0.232 & 0.242 \\
    Gaussian                     & 0.260 & 0.264 & 0.200 & 0.197 \\
    Gaussian ING                 & 0.264 & 0.263 & 0.201 & 0.189 \\
    Bernoulli Diversity          & 0.305 & \textbf{0.371} & 0.237 & 0.247 \\
    Bernoulli Diversity Concave  & 0.306 & \textbf{0.371} & 0.236 & 0.247 \\
    RL Top-$k$ UCB diversity     & 0.324 & \textbf{0.390} & 0.256 & 0.249 \\
    \bottomrule
  \end{tabular}
  }
  \caption{Comparison of precision under budgets \(b = 10\%\) and \(b=20\%\) for both serialized (Ser.) and hierarchical (Hier.) document selection strategies with correlated bandits. For all discrete estimate policies, selecting a subset of documents by taking into account sub-query hierarchies yields better performance, with Bernoulli top \(k=5\) having a \(24\%\) performance boost over the serialized representation, where we use the number of observations \(n=4\), informativeness threshold \(\tau=0.77\) and inheritance factor \(\lambda=0.91\)}
  \label{tab:budget10_results}
\end{table}

\subsection{NeuCLIR analysis}\label{section:neuclir_analysis}

\begin{figure}[ht]
    \centering
    \includegraphics[clip,trim=12mm 4mm 12mm 12mm, width=\linewidth]{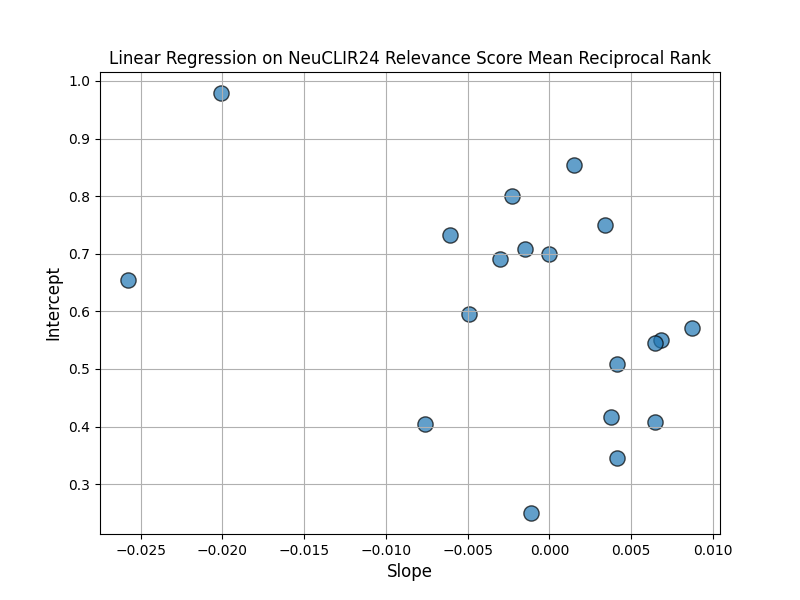}
    \caption{Fitting linear regression functions \(y = \alpha + \beta x +\epsilon\) between the mean reciprocal ranks \(\bar{r}_i = \frac{1}{K}\sum_{k=1}^K r_{i,k}\) for each user request in NeuCLIR24 (represented by a point).}
    \label{fig:slope_neuclir}
\end{figure}

\header{Rank information} We run our experiments in a setting where each sub-query corresponds to a ranked list of documents. Therefore, we make the assumption that the ranked list is a good reflection of document relevance. To verify this hypothesis, we calculate the mean reciprocal rank and fit a linear regression to estimate whether the rank is negatively correlated with the relevance of the documents. Figure \ref{fig:slope_neuclir} shows that very few instances have a negative slope, indicating that the ranked lists are not always consistent with relevance.

\begin{table}[ht]
  \centering
  \setlength{\tabcolsep}{3.5pt}
  \renewcommand{\arraystretch}{1.2}
  \resizebox{\linewidth}{!}{%
  \begin{tabular}{l *{5}{c} *{2}{c}}
    \toprule
     & \multicolumn{5}{c}{\makecell{\textbf{\(\alpha\text{-nDCG}\)} $\uparrow$ \\ NeuCLIR24}} 
     & \multicolumn{2}{c}{\makecell{\textbf{\(\alpha\text{-nDCG}\)} $\uparrow$ \\ Researchy}} \\
    \cmidrule(lr){2-6} \cmidrule(lr){7-8}
     & $K{=}5$ & $K{=}10$ & $K{=}20$ & $K{=}40$ & $K{=}50$ & $K{=}5$ & $K{=}10$ \\
    \midrule
    Bernoulli            & 0.445 & 0.508 & 0.543 & 0.246 & 0.282 & 0.210 & \textbf{0.157} \\
    \midrule
    Gaussian             & 0.413 & 0.442 & 0.468 & 0.252 & 0.281 & 0.172 & 0.140 \\
    Gaussian ING         & 0.409 & 0.445 & 0.463 & 0.259 & 0.284 & 0.170 & 0.135 \\
    \midrule
    Bernoulli $k=3$      & \textbf{0.462} & 0.531 & \textbf{0.556} & 0.264 & 0.285 & \textbf{0.230} & 0.155 \\
    Bernoulli $k=4$      & 0.461 & \textbf{0.536} & 0.551 & \textbf{0.275} & 0.283 & \textbf{0.240} & 0.153 \\
    Bernoulli $k=5$      & 0.448 & 0.523 & \textbf{0.555} & \textbf{0.267} & \textbf{0.291} & \textbf{0.232} & \textbf{0.156} \\
    \bottomrule
  \end{tabular}
  }
  \caption{Comparison of $\alpha$-nDCG for NeuCLIR24 and ResearcyQuestions. Gaussian vs.\ Bernoulli modelling.}
  \label{tab:Gaussian_alpha_ndcg}
\end{table}

\header{Top-$k$} We explore different values of the top \(k\) documents in a ranked list. More precisely, we investigate performance for values of \(k \in [3, 4, 5]\). We observed that any Bernoulli taking into account top-$k$ documents outperforms all the other rewards across budget, and therefore we report only top-$k$ for $k = 3$ in the main tables. In Table~\ref{tab:Gaussian_alpha_ndcg} we report the values for all experiments as reference. For analysis on Gaussian distributions, check \ref{sec:gaussian}.

\section{Conclusions}

This study proposes reframing query decomposition and document retrieval for complex query answering as an exploitation–exploration problem under a multi-armed bandit setting. We conduct extensive empirical studies across a range of policies and discover that, when having access to binary relevance labels, Bernoulli distributions with rank information perform best for selecting relevant content under a constrained budget. These results are demonstrated on two datasets, NeuCLIR24 and ResearchyQuestions. We further show that modeling correlations between sub-queries in a hierarchical setting consistently boosts performance across all rewards. We also stress that such an approach performs robustly, while the choice of budget is data-dependent w.r.t.\ the real distribution of relevance across retrieved documents.

We propose further analyses into correlated and contextualized bandits where correlations need not be explicit in the data but may instead emerge naturally in the representational space of sub-queries. Another direction would be to train a retrieval model optimized for relevance, document diversity, exploration, or long-form generation within a policy-learning framework.

\clearpage
\section*{Limitations}
Our work proposes a reframing of an existing paradigm in a popular experimental setup. The study comes with a couple of challenges and limitations: when a retrieval module is involved, the rewards directly depend on the performance of the retriever; moreover, the performance over budgets directly depends on the real distribution of observations. In a dynamic setting, often a set of sub-queries is not pre-defined, such as in the NeuCLIR dataset, which introduces noise by regenerating sub-queries every time. In addition, often the sub-queries are not only not generated, but they are samples from an unbounded search space for a given user request.

\section*{Acknowledgments}
This research was (partially) supported by the Dutch Research Council (NWO), under project numbers 024.004.022, NWA.1389.20.\-183, and KICH3.LTP.20.006, and the European Union under grant agreements No. 101070212 (FINDHR) and No. 101201510 (UNITE).

Views and opinions expressed are those of the authors only and do not necessarily reflect those of their respective employers, funders and/or granting authorities.

\bibliography{references}

\begin{thebibliography}{37}
\providecommand{\natexlab}[1]{#1}

\bibitem[{Asai et~al.(2020)Asai, Hashimoto, Hajishirzi, Socher, and Xiong}]{asai2020multipledocuments}
Akari Asai, Kazuma Hashimoto, Hannaneh Hajishirzi, Richard Socher, and Caiming Xiong. 2020.
\newblock \href {https://openreview.net/forum?id=SJgVHkrYDH} {Learning to retrieve reasoning paths over wikipedia graph for question answering}.
\newblock In \emph{8th International Conference on Learning Representations, {ICLR} 2020, Addis Ababa, Ethiopia, April 26-30, 2020}. OpenReview.net.

\bibitem[{Askari et~al.(2025)Askari, Petcu, Meng, Aliannejadi, Abolghasemi, Kanoulas, and Verberne}]{askari2025selfseedingmultiintentselfinstructingllms}
Arian Askari, Roxana Petcu, Chuan Meng, Mohammad Aliannejadi, Amin Abolghasemi, Evangelos Kanoulas, and Suzan Verberne. 2025.
\newblock \href {https://arxiv.org/abs/2402.11633} {Self-seeding and multi-intent self-instructing llms for generating intent-aware information-seeking dialogs}.
\newblock \emph{Preprint}, arXiv:2402.11633.

\bibitem[{Bai et~al.(2023)Bai, Bai, Chu, Cui, Dang, Deng, Fan, Ge, Han, Huang, Hui, Ji, Li, Lin, Lin, Liu, Liu, Lu, Lu, Ma, Men, Ren, Ren, Tan, Tan, Tu, Wang, Wang, Wang, Wu, Xu, Xu, Yang, Yang, Yang, Yang, Yao, Yu, Yuan, Yuan, Zhang, Zhang, Zhang, Zhang, Zhou, Zhou, Zhou, and Zhu}]{bai2023qwentechnicalreport}
Jinze Bai, Shuai Bai, Yunfei Chu, Zeyu Cui, Kai Dang, Xiaodong Deng, Yang Fan, Wenbin Ge, Yu~Han, Fei Huang, Binyuan Hui, Luo Ji, Mei Li, Junyang Lin, Runji Lin, Dayiheng Liu, Gao Liu, Chengqiang Lu, Keming Lu, and 29 others. 2023.
\newblock \href {https://arxiv.org/abs/2309.16609} {Qwen technical report}.
\newblock \emph{Preprint}, arXiv:2309.16609.

\bibitem[{Bhargav et~al.(2022)Bhargav, Sidiropoulos, and Kanoulas}]{tip_of_the_tongue}
Samarth Bhargav, Georgios Sidiropoulos, and Evangelos Kanoulas. 2022.
\newblock \href {https://doi.org/10.1145/3488560.3498421} {`{It}'s on the tip of my tongue': {A} new dataset for known-item retrieval}.
\newblock In \emph{{WSDM} '22: The Fifteenth {ACM} International Conference on Web Search and Data Mining, Virtual Event / Tempe, AZ, USA, February 21 - 25, 2022}, pages 48--56. {ACM}.

\bibitem[{Buck et~al.(2018)Buck, Bulian, Ciaramita, Gajewski, Gesmundo, Houlsby, and Wang}]{Buck2018ReformulationQuery}
Christian Buck, Jannis Bulian, Massimiliano Ciaramita, Wojciech Gajewski, Andrea Gesmundo, Neil Houlsby, and Wei Wang. 2018.
\newblock \href {https://openreview.net/forum?id=S1CChZ-CZ} {Ask the right questions: Active question reformulation with reinforcement learning}.
\newblock In \emph{6th International Conference on Learning Representations, {ICLR} 2018, Vancouver, BC, Canada, April 30 - May 3, 2018, Conference Track Proceedings}. OpenReview.net.

\bibitem[{Clarke et~al.(2008)Clarke, Kolla, Cormack, Vechtomova, Ashkan, B\"{u}ttcher, and MacKinnon}]{alphandcg}
Charles~L.A. Clarke, Maheedhar Kolla, Gordon~V. Cormack, Olga Vechtomova, Azin Ashkan, Stefan B\"{u}ttcher, and Ian MacKinnon. 2008.
\newblock \href {https://doi.org/10.1145/1390334.1390446} {Novelty and diversity in information retrieval evaluation}.
\newblock In \emph{Proceedings of the 31st Annual International ACM SIGIR Conference on Research and Development in Information Retrieval}, SIGIR '08, page 659–666, New York, NY, USA. Association for Computing Machinery.

\bibitem[{Deng et~al.(2022)Deng, Wang, Hsieh, Wang, Guo, Shu, Song, Xing, and Hu}]{deng2022rlpromptoptimizingdiscretetext}
Mingkai Deng, Jianyu Wang, Cheng-Ping Hsieh, Yihan Wang, Han Guo, Tianmin Shu, Meng Song, Eric~P. Xing, and Zhiting Hu. 2022.
\newblock \href {https://arxiv.org/abs/2205.12548} {Rlprompt: Optimizing discrete text prompts with reinforcement learning}.
\newblock \emph{Preprint}, arXiv:2205.12548.

\bibitem[{Gu et~al.(2025)Gu, Jia, Piccardi, and Yu}]{Gu2025ChainOfRetrieval}
Zhanzhong Gu, Wenjing Jia, Massimo Piccardi, and Ping Yu. 2025.
\newblock \href {https://doi.org/10.1016/J.ARTMED.2025.103078} {Empowering large language models for automated clinical assessment with generation-augmented retrieval and hierarchical chain-of-thought}.
\newblock \emph{Artif. Intell. Medicine}, 162:103078.

\bibitem[{Jin et~al.(2024)Jin, Yoon, Han, and Arik}]{jin2024longcontextllmsmeetrag}
Bowen Jin, Jinsung Yoon, Jiawei Han, and Sercan~O. Arik. 2024.
\newblock \href {https://arxiv.org/abs/2410.05983} {Long-context llms meet rag: Overcoming challenges for long inputs in rag}.
\newblock \emph{Preprint}, arXiv:2410.05983.

\bibitem[{Jin et~al.(2025)Jin, Zeng, Yue, Yoon, Arik, Wang, Zamani, and Han}]{Jun2025SearchR1}
Bowen Jin, Hansi Zeng, Zhenrui Yue, Jinsung Yoon, Sercan Arik, Dong Wang, Hamed Zamani, and Jiawei Han. 2025.
\newblock \href {https://arxiv.org/abs/2503.09516} {Search-r1: Training llms to reason and leverage search engines with reinforcement learning}.
\newblock \emph{Preprint}, arXiv:2503.09516.

\bibitem[{Khot et~al.(2023)Khot, Trivedi, Finlayson, Fu, Richardson, Clark, and Sabharwal}]{khot2023decomposedpromptingmodularapproach}
Tushar Khot, Harsh Trivedi, Matthew Finlayson, Yao Fu, Kyle Richardson, Peter Clark, and Ashish Sabharwal. 2023.
\newblock \href {https://arxiv.org/abs/2210.02406} {Decomposed prompting: A modular approach for solving complex tasks}.
\newblock \emph{Preprint}, arXiv:2210.02406.

\bibitem[{Kim et~al.(2025)Kim, Lee, Jang, Cho, Kim, Cho, and Lee}]{kim2025NotInformative}
Takyoung Kim, Kyungjae Lee, Young~Rok Jang, Ji~Yong Cho, Gangwoo Kim, Minseok Cho, and Moontae Lee. 2025.
\newblock \href {https://arxiv.org/abs/2407.01158} {Learning to explore and select for coverage-conditioned retrieval-augmented generation}.
\newblock \emph{Preprint}, arXiv:2407.01158.

\bibitem[{Krasakis et~al.(2025)Krasakis, Yates, and Kanoulas}]{krasakis2025constructingsetcompositionalnegatedrepresentations}
Antonios~Minas Krasakis, Andrew Yates, and Evangelos Kanoulas. 2025.
\newblock \href {https://arxiv.org/abs/2501.07679} {Constructing set-compositional and negated representations for first-stage ranking}.
\newblock \emph{Preprint}, arXiv:2501.07679.

\bibitem[{Lawrie et~al.(2025)Lawrie, MacAvaney, Mayfield, McNamee, Oard, Soldaini, and Yang}]{lawrie2025overviewtrec2024neuclir}
Dawn Lawrie, Sean MacAvaney, James Mayfield, Paul McNamee, Douglas~W. Oard, Luca Soldaini, and Eugene Yang. 2025.
\newblock \href {https://arxiv.org/abs/2509.14355} {Overview of the trec 2024 neuclir track}.
\newblock \emph{Preprint}, arXiv:2509.14355.

\bibitem[{Lewis et~al.(2021)Lewis, Perez, Piktus, Petroni, Karpukhin, Goyal, Küttler, Lewis, tau Yih, Rocktäschel, Riedel, and Kiela}]{lewis2021retrievalaugmentedgenerationknowledgeintensivenlp}
Patrick Lewis, Ethan Perez, Aleksandra Piktus, Fabio Petroni, Vladimir Karpukhin, Naman Goyal, Heinrich Küttler, Mike Lewis, Wen tau Yih, Tim Rocktäschel, Sebastian Riedel, and Douwe Kiela. 2021.
\newblock \href {https://arxiv.org/abs/2005.11401} {Retrieval-augmented generation for knowledge-intensive nlp tasks}.
\newblock \emph{Preprint}, arXiv:2005.11401.

\bibitem[{Li and Kanoulas(2017)}]{Li_2017}
Dan Li and Evangelos Kanoulas. 2017.
\newblock \href {https://doi.org/10.1145/3132847.3133015} {Active sampling for large-scale information retrieval evaluation}.
\newblock In \emph{Proceedings of the 2017 ACM on Conference on Information and Knowledge Management}, CIKM ’17, page 49–58. ACM.

\bibitem[{Nair et~al.(2022)Nair, Yang, Lawrie, Duh, McNamee, Murray, Mayfield, and Oard}]{colbert-x}
Suraj Nair, Eugene Yang, Dawn Lawrie, Kevin Duh, Paul McNamee, Kenton Murray, James Mayfield, and Douglas~W. Oard. 2022.
\newblock \href {https://arxiv.org/abs/2201.08471} {Transfer learning approaches for building cross-language dense retrieval models}.
\newblock In \emph{Proceedings of the 44th European Conference on Information Retrieval (ECIR)}.

\bibitem[{Nguyen et~al.(2023)Nguyen, MacAvaney, and Yates}]{LSRandrewThong}
Thong Nguyen, Sean MacAvaney, and Andrew Yates. 2023.
\newblock \href {https://arxiv.org/abs/2303.13416} {A unified framework for learned sparse retrieval}.
\newblock \emph{Preprint}, arXiv:2303.13416.

\bibitem[{Odijk et~al.(2015)Odijk, Meij, Sijaranamual, and de~Rijke}]{odijkMaarten}
Daan Odijk, Edgar Meij, Isaac Sijaranamual, and Maarten de~Rijke. 2015.
\newblock \href {https://doi.org/10.1145/2766462.2767715} {Dynamic query modeling for related content finding}.
\newblock In \emph{Proceedings of the 38th International {ACM} {SIGIR} Conference on Research and Development in Information Retrieval, Santiago, Chile, August 9-13, 2015}, pages 33--42. {ACM}.

\bibitem[{Petcu et~al.(2025)Petcu, Bhargav, de~Rijke, and Kanoulas}]{petcu2025comprehensivetaxonomynegationnlp}
Roxana Petcu, Samarth Bhargav, Maarten de~Rijke, and Evangelos Kanoulas. 2025.
\newblock \href {https://arxiv.org/abs/2507.22337} {A comprehensive taxonomy of negation for nlp and neural retrievers}.
\newblock \emph{Preprint}, arXiv:2507.22337.

\bibitem[{Petcu and Maji(2024)}]{petcu2024efficientdataselectionemploying}
Roxana Petcu and Subhadeep Maji. 2024.
\newblock \href {https://arxiv.org/abs/2402.14888} {Efficient data selection employing semantic similarity-based graph structures for model training}.
\newblock \emph{Preprint}, arXiv:2402.14888.

\bibitem[{Podolak et~al.(2025)Podolak, Peri\'{c}, Jani\'{c}ijevi\'{c}, and Petcu}]{SetwiseInsertion2024}
Jakub Podolak, Leon Peri\'{c}, Mina Jani\'{c}ijevi\'{c}, and Roxana Petcu. 2025.
\newblock \href {https://doi.org/10.1145/3726302.3730323} {Beyond reproducibility: Advancing zero-shot llm reranking efficiency with setwise insertion}.
\newblock In \emph{Proceedings of the 48th International ACM SIGIR Conference on Research and Development in Information Retrieval}, SIGIR '25, page 3205–3213, New York, NY, USA. Association for Computing Machinery.

\bibitem[{Qi et~al.(2019)Qi, Lin, Mehr, Wang, and Manning}]{reasoning_missing_entities}
Peng Qi, Xiaowen Lin, Leo Mehr, Zijian Wang, and Christopher~D. Manning. 2019.
\newblock \href {https://doi.org/10.18653/V1/D19-1261} {Answering complex open-domain questions through iterative query generation}.
\newblock In \emph{Proceedings of the 2019 Conference on Empirical Methods in Natural Language Processing and the 9th International Joint Conference on Natural Language Processing, {EMNLP-IJCNLP} 2019, Hong Kong, China, November 3-7, 2019}, pages 2590--2602. Association for Computational Linguistics.

\bibitem[{Rafailov et~al.(2024)Rafailov, Sharma, Mitchell, Ermon, Manning, and Finn}]{rafailov2024DPO}
Rafael Rafailov, Archit Sharma, Eric Mitchell, Stefano Ermon, Christopher~D. Manning, and Chelsea Finn. 2024.
\newblock \href {https://arxiv.org/abs/2305.18290} {Direct preference optimization: Your language model is secretly a reward model}.
\newblock \emph{Preprint}, arXiv:2305.18290.

\bibitem[{Rahman et~al.(2020)Rahman, Kutlu, Elsayed, and Lease}]{Rahman_2020}
Md~Mustafizur Rahman, Mucahid Kutlu, Tamer Elsayed, and Matthew Lease. 2020.
\newblock \href {https://doi.org/10.1145/3409256.3409837} {Efficient test collection construction via active learning}.
\newblock In \emph{Proceedings of the 2020 ACM SIGIR on International Conference on Theory of Information Retrieval}, ICTIR ’20, page 177–184. ACM.

\bibitem[{Rosset et~al.(2024)Rosset, Chung, Qin, Chau, Feng, Awadallah, Neville, and Rao}]{rosset2024researchyquestionsdatasetmultiperspective}
Corby Rosset, Ho-Lam Chung, Guanghui Qin, Ethan~C. Chau, Zhuo Feng, Ahmed Awadallah, Jennifer Neville, and Nikhil Rao. 2024.
\newblock \href {https://arxiv.org/abs/2402.17896} {Researchy questions: A dataset of multi-perspective, decompositional questions for llm web agents}.
\newblock \emph{Preprint}, arXiv:2402.17896.

\bibitem[{Shao et~al.(2025)Shao, Qiao, Kishore, Muennighoff, Lin, Rus, Low, Min, tau Yih, Koh, and Zettlemoyer}]{shao2025reasonirtrainingretrieversreasoning}
Rulin Shao, Rui Qiao, Varsha Kishore, Niklas Muennighoff, Xi~Victoria Lin, Daniela Rus, Bryan Kian~Hsiang Low, Sewon Min, Wen tau Yih, Pang~Wei Koh, and Luke Zettlemoyer. 2025.
\newblock \href {https://arxiv.org/abs/2504.20595} {Reasonir: Training retrievers for reasoning tasks}.
\newblock \emph{Preprint}, arXiv:2504.20595.

\bibitem[{Soudani et~al.(2024)Soudani, Petcu, Kanoulas, and Hasibi}]{soudani2024surveyrecentadvancesconversational}
Heydar Soudani, Roxana Petcu, Evangelos Kanoulas, and Faegheh Hasibi. 2024.
\newblock \href {https://arxiv.org/abs/2405.13003} {A survey on recent advances in conversational data generation}.
\newblock \emph{Preprint}, arXiv:2405.13003.

\bibitem[{van~den Elsen et~al.(2025)van~den Elsen, Barkhof, Nijdam, Lupart, and Aliannejadi}]{repro_nevir}
Coen van~den Elsen, Francien Barkhof, Thijmen Nijdam, Simon Lupart, and Mohammad Aliannejadi. 2025.
\newblock \href {https://doi.org/10.1145/3726302.3730294} {Reproducing nevir: Negation in neural information retrieval}.
\newblock In \emph{Proceedings of the 48th International {ACM} {SIGIR} Conference on Research and Development in Information Retrieval, {SIGIR} 2025, Padua, Italy, July 13-18, 2025}, pages 3346--3356. {ACM}.

\bibitem[{Voorhees(2018)}]{10.1145/3269206.3271766}
Ellen~M. Voorhees. 2018.
\newblock \href {https://doi.org/10.1145/3269206.3271766} {On building fair and reusable test collections using bandit techniques}.
\newblock In \emph{Proceedings of the 27th ACM International Conference on Information and Knowledge Management}, CIKM '18, page 407–416, New York, NY, USA. Association for Computing Machinery.

\bibitem[{Walden et~al.(2025)Walden, Mason, Weller, Dietz, Recknor, Li, Liu, Hou, Mayfield, and Yang}]{autoargue}
William Walden, Marc Mason, Orion Weller, Laura Dietz, Hannah Recknor, Bryan Li, Gabrielle Kaili-May Liu, Yu~Hou, James Mayfield, and Eugene Yang. 2025.
\newblock \href {https://arxiv.org/abs/2509.26184} {Auto-argue: Llm-based report generation evaluation}.
\newblock \emph{Preprint}, arXiv:2509.26184.

\bibitem[{Weller et~al.(2024)Weller, Lawrie, and Durme}]{nevir}
Orion Weller, Dawn~J. Lawrie, and Benjamin~Van Durme. 2024.
\newblock \href {https://aclanthology.org/2024.eacl-long.139} {Nevir: Negation in neural information retrieval}.
\newblock In \emph{Proceedings of the 18th Conference of the European Chapter of the Association for Computational Linguistics, {EACL} 2024 - Volume 1: Long Papers, St. Julian's, Malta, March 17-22, 2024}, pages 2274--2287. Association for Computational Linguistics.

\bibitem[{Yang et~al.(2025)Yang, Yang, Zhang, Hui, Zheng, Yu, Li, Liu, Huang, Wei, Lin, Yang, Tu, Zhang, Yang, Yang, Zhou, Lin, Dang, Lu, Bao, Yang, Yu, Li, Xue, Zhang, Zhu, Men, Lin, Li, Tang, Xia, Ren, Ren, Fan, Su, Zhang, Wan, Liu, Cui, Zhang, and Qiu}]{qwen2025qwen25technicalreport}
An~Yang, Baosong Yang, Beichen Zhang, Binyuan Hui, Bo~Zheng, Bowen Yu, Chengyuan Li, Dayiheng Liu, Fei Huang, Haoran Wei, Huan Lin, Jian Yang, Jianhong Tu, Jianwei Zhang, Jianxin Yang, Jiaxi Yang, Jingren Zhou, Junyang Lin, Kai Dang, and 23 others. 2025.
\newblock \href {https://arxiv.org/abs/2412.15115} {Qwen2.5 technical report}.
\newblock \emph{Preprint}, arXiv:2412.15115.

\bibitem[{Yang et~al.(2024)Yang, Lawrie, Mayfield, Oard, and Miller}]{translate-distill}
Eugene Yang, Dawn Lawrie, James Mayfield, Douglas~W. Oard, and Scott Miller. 2024.
\newblock \href {https://arxiv.org/abs/2401.04810} {Translate-distill: Learning cross-language dense retrieval by translation and distillation}.
\newblock In \emph{Proceedings of the 46th European Conference on Information Retrieval (ECIR)}.

\bibitem[{Yao et~al.(2023)Yao, Zhao, Yu, Du, Shafran, Narasimhan, and Cao}]{yaoreasoningpaths}
Shunyu Yao, Jeffrey Zhao, Dian Yu, Nan Du, Izhak Shafran, Karthik~R. Narasimhan, and Yuan Cao. 2023.
\newblock \href {https://openreview.net/forum?id=WE\_vluYUL-X} {React: Synergizing reasoning and acting in language models}.
\newblock In \emph{The Eleventh International Conference on Learning Representations, {ICLR} 2023, Kigali, Rwanda, May 1-5, 2023}. OpenReview.net.

\bibitem[{Zhang et~al.(2025)Zhang, Zhang, Wu, Pei, Ren, de~Rijke, Chen, and Ren}]{excluir}
Wenhao Zhang, Mengqi Zhang, Shiguang Wu, Jiahuan Pei, Zhaochun Ren, Maarten de~Rijke, Zhumin Chen, and Pengjie Ren. 2025.
\newblock \href {https://doi.org/10.1609/AAAI.V39I12.33451} {Excluir: Exclusionary neural information retrieval}.
\newblock In \emph{AAAI-25, Sponsored by the Association for the Advancement of Artificial Intelligence, February 25 - March 4, 2025, Philadelphia, PA, {USA}}, pages 13295--13303. {AAAI} Press.

\bibitem[{Zhang et~al.(2024)Zhang, Zhu, gang Zhou, Qi, Zhang, and Li}]{Zhang2024BoolQuestionsDD}
Zongmeng Zhang, Jinhua Zhu, Wen gang Zhou, Xiang Qi, Peng Zhang, and Houqiang Li. 2024.
\newblock {BoolQuestions}: Does dense retrieval understand boolean logic in language?
\newblock In \emph{Conference on Empirical Methods in Natural Language Processing}.

\end{thebibliography}

\appendix
\section{Appendix}

\subsection{Prompts for Query Decomposition}\label{sec:prompts}
The NeuCLIR dataset contains complex user requests and associated nuggets with annotated documents. We thus generate our own query decomposition using LLM calls. We decompose each NeuCLIR user request into \(k=16\) sub-queries using prompts \ref{fig:problem-definition-prompt} and \ref{fig:specialized-information-prompt}.

\begin{figure*}[!ht]
  \centering
    \begin{subfigure}{\linewidth}
	\begin{tcolorbox}[colback=gray!5!white, colframe=black!75!black, boxrule=0.5pt, arc=3pt, left=6pt, right=6pt, top=4pt, bottom=4pt]
	
	\textbf{Prompt for generating Problem Definition Search Queries}

	You are an assistant that receives a complex user request and must generate search queries whose answers:
	(1) define the user's request, and 
	(2) define all key entities mentioned in the request.
	
	\begin{enumerate}
		\item Consider any current events, recent developments, or specific details mentioned in the context to enhance the queries (if context is provided).
		\item Write \textbf{up to \texttt{<<no\_queries>>}} objective Google search queries related to the request.\\
		      The queries should be general and high-level (e.g., encyclopedic or definitional in nature).\\
		      Avoid narrow or overly specific phrasings. Queries should be phrased as something someone would realistically search for.
		\item Assume the current date is \texttt{<<current\_date>>} if required by the request.
		\item Return the queries as a JSON list of strings in the format: \\
		      \texttt{["query 1", "query 2", "query 3"]}.\\
		      The response must contain \textbf{only} this list.
	\end{enumerate}
    \end{tcolorbox}
    \end{subfigure}
  \caption{Prompt for Generating Problem Definition Queries}
  \label{fig:problem-definition-prompt}
\end{figure*}

\begin{figure*}[!ht]
  \centering
    \begin{subfigure}{\linewidth}
    \begin{tcolorbox}[colback=gray!5!white, colframe=black!75!black, boxrule=0.5pt, arc=3pt, left=6pt, right=6pt, top=4pt, bottom=4pt]
    
    \textbf{Prompt for Generating Specialized Information Search Queries}

    You are a seasoned research assistant tasked with generating search queries that ask highly specialized and detailed information about a user request. You have access to a context formed of multiple supporting documents.

    \begin{enumerate}
        \item Use the context (if provided) to identify aspects of the request that would benefit from more specific, in-depth, or expert-level information.\\
              Consider any current events, real-time developments, or precise facts that can be used to craft better questions.
        \item Write \textbf{up to \texttt{<<no\_queries>>}} objective Google search queries related to the request.\\
              These queries should be highly specific and targeted—something a person wouldn't know from general knowledge.\\
              Each query should reflect a concrete sub-aspect or follow-up inquiry about the original request.
        \item Assume the current date is \texttt{<<current\_date>>} if required by the request.
        \item Return the queries as a JSON list of strings in the format: \\
              \texttt{["query 1", "query 2", "query 3"]}.\\
              The response must contain \textbf{only} this list.
    \end{enumerate}
    \end{tcolorbox}
    \end{subfigure}
  \caption{Prompt for Generating Specialized Information Queries}
  \label{fig:specialized-information-prompt}
\end{figure*}

\subsection{Gaussian Distributions}\label{sec:gaussian}

\header{Gaussian estimates} The finding in Figure \ref{fig:slope_neuclir} serves as an explanation to the Gaussian policies performing considerably poorly compared to the Bernoulli ones, as seen Table \ref{tab:Gaussian_alpha_ndcg}. This suggests that Gaussian distributions may not be suited to capturing relevance when the ranking signal is noisy. 

\begin{algorithm}[ht]
\caption{Thompson Sampling in Continuous Space}\label{tab:thompson_continuous_space}
\SetKwInOut{Input}{Input}
\SetKwInOut{Output}{Output}
\Input{Sub-queries $\mathcal{S} = \{s_1, \dots, s_K\}$}
\Input{Budget $b$}
\BlankLine

\For{$i = 1,\dots,K$}{
    Define Gaussian priors $\mathcal{N}(\mu_0, \sigma_0^2)$ where $\mu_0 \gets 0$ and $\sigma_0^2 \gets 1$ \\
}

\For{$t = 1,\dots,b$}{
    \For{$i = 1,\dots,K$}{
        Sample from posterior $\theta_i \sim \mathcal{N}(\mu_i,\sigma^2_i)$
    }
    Select arm (sub-query): $a_t \gets \arg\max_{i} \theta_i$ \\
    
    Pick next document $d_{a_t, n} \in \mathcal{D}_{a_t}$ where \\
    \hspace{1em} \(n = \min \{ j \in \{1,\ldots,N\} \mid (a_t, j) \notin \mathcal{O} \}\)
    
    Observe the likelihood: $r(a_t, n)$

    Update posterior: \\
       \hspace{1em} $\sigma_{a_t}^2 = \sqrt{(\frac{1}{\sigma_0^2} + n)^{-1}}$ \\
       \hspace{1em} $\mu_{a_t} = \sigma_{a_t}^2 \cdot \sum_{r \in {S_{a_t}}} r$
}
\end{algorithm}

\subsection{Report Generation Example}\label{sec:app_report_gen_example}
In this section, we show an example of a report generated and evaluated using Auto-ARGUE.

\header{Query} What characterizes Ernest Hemingway’s writing style, and what is the “iceberg theory”?

\header{Generated report} Ernest Hemingway was a renowned American writer known for a distinctive style built on simplicity, clarity, and concrete detail \([5]\).
He is often associated with the “iceberg theory,” the idea that most meaning lies beneath the surface while only a small portion is stated explicitly \([3],[5]\).
Hemingway’s years as a journalist shaped this approach: he learned to write concise, unadorned prose that focuses on essentials \([2]\).
He also insisted on authenticity—“never write about what you do not know”—and favored describing observable actions while letting readers infer the emotions behind them \([1]\).
This restrained, lucid style has been widely emulated and is a major reason he is regarded as one of the most influential writers of the twentieth century \([4]\).

\subsection{Complete results on NeuCLIR and ResearchyQuestions}

\begin{figure*}[t]
    \centering
    \includegraphics[width=\linewidth]{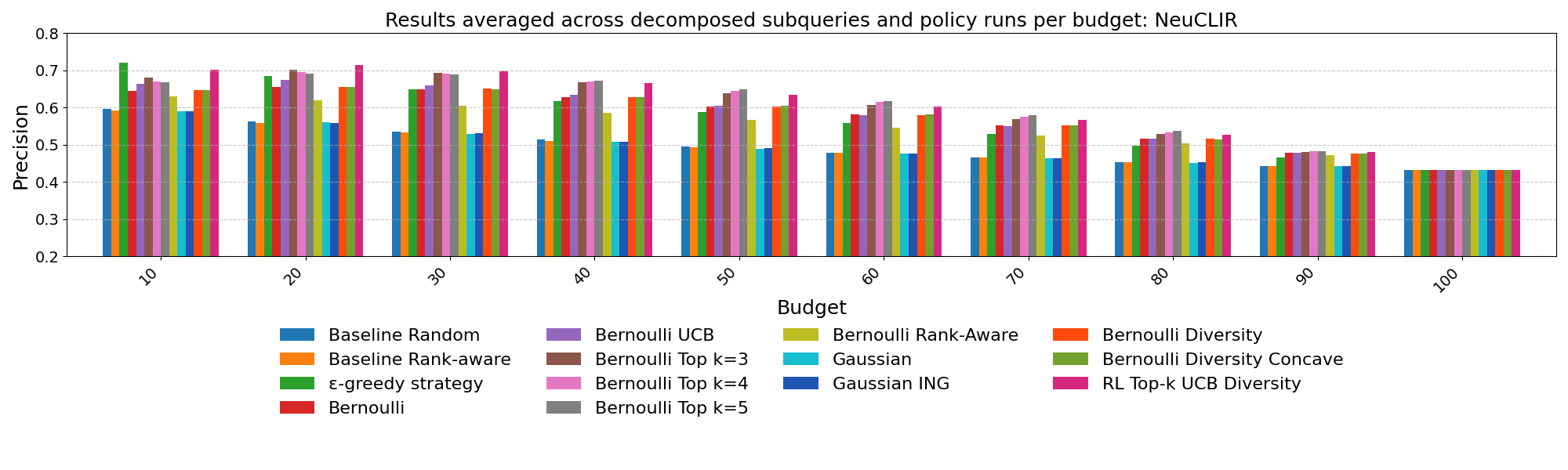}
    \caption{Precision applying baselines and reward policies for query selection for different budgets over observed documents. The budget is calculated as a percentage over the total amount of available documents. Modelling the queries using Bernoulli distribution considering the top ranked k documents performs the best across budgets, with all policies reaching the same performance as the budget covers the whole space of actions.}
    \label{fig:neuclir_precision_all}
\end{figure*}

\begin{figure*}[t]
    \centering
    \includegraphics[width=\linewidth]{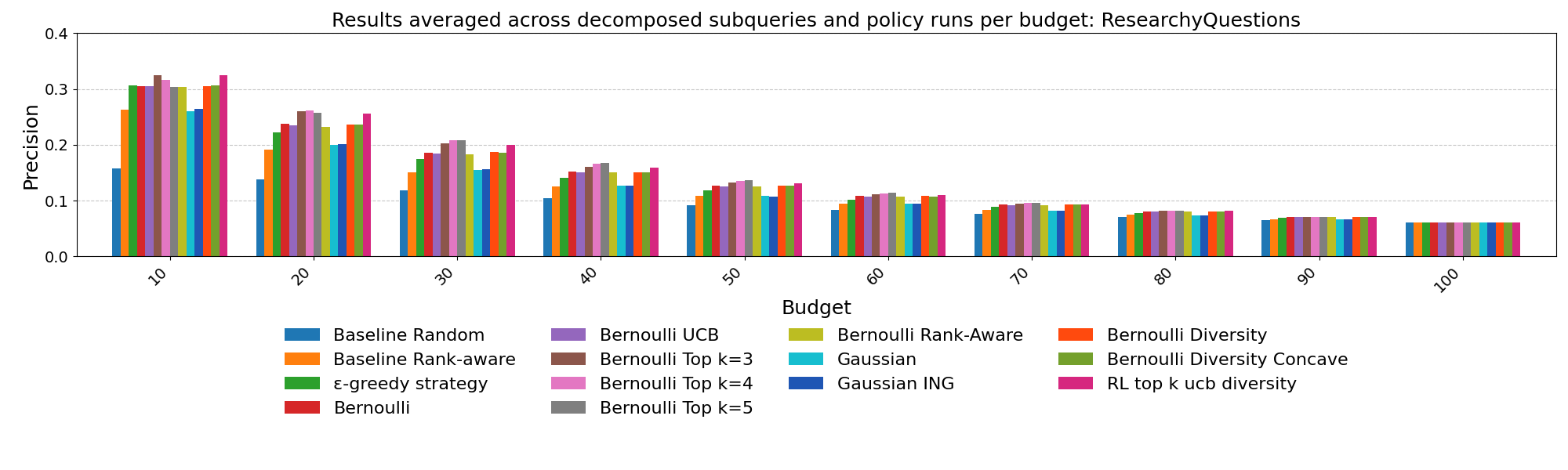}
    \caption{Precision applying baselines and reward policies for query selection for different budgets over observed documents in ResearchyQuestions. The budget is calculated as a percentage over the total amount of available documents. Modeling the queries using Bernoulli distribution considering the top ranked k documents performs the best across budgets, with all policies reaching the same performance as the budget covers the whole space of actions.}
    \label{fig:researchy_eval_all}
\end{figure*}

\begin{figure*}[t]
    \centering
    \includegraphics[width=\linewidth]{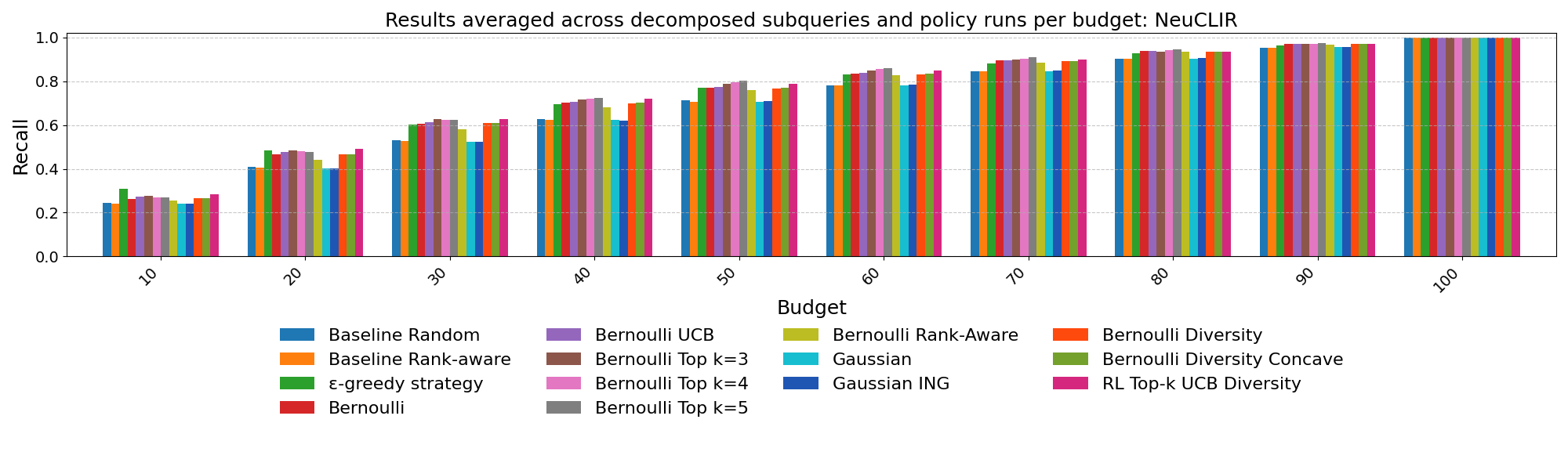}
    \caption{Recall applying baselines and reward policies for query selection for different budgets over observed documents. The budget is calculated as a percentage over the total amount of available documents. Modelling the queries using Bernoulli distribution considering the top ranked k documents performs the best across budgets, with all policies reaching the same performance as the budget covers the whole space of actions.}
    \label{fig:neuclir_recall_all}
\end{figure*}

\begin{figure*}[t]
    \centering
    \includegraphics[width=\linewidth]{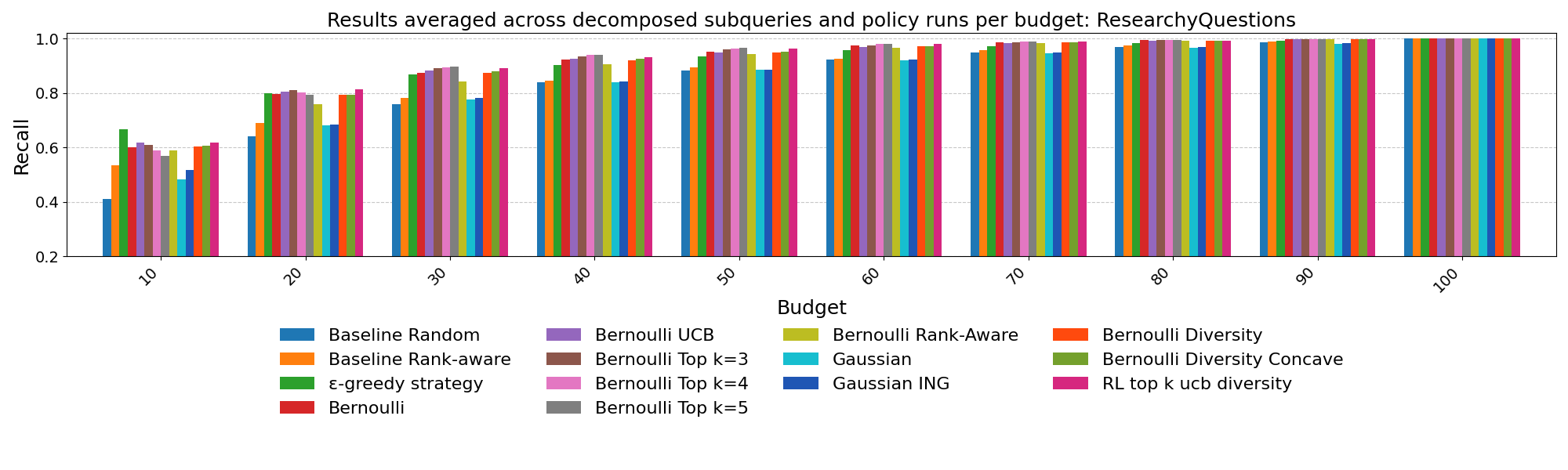}
    \caption{Recall applying baselines and reward policies for query selection for different budgets over observed documents in ResearchyQuestions. The budget is calculated as a percentage over the total amount of available documents. Modeling the queries using Bernoulli distribution considering the top ranked k documents performs the best across budgets, with all policies reaching the same performance as the budget covers the whole space of actions.}
    \label{fig:researchy_recall_all}
\end{figure*}

\begin{figure*}[t]
    \centering
    \includegraphics[width=\linewidth]{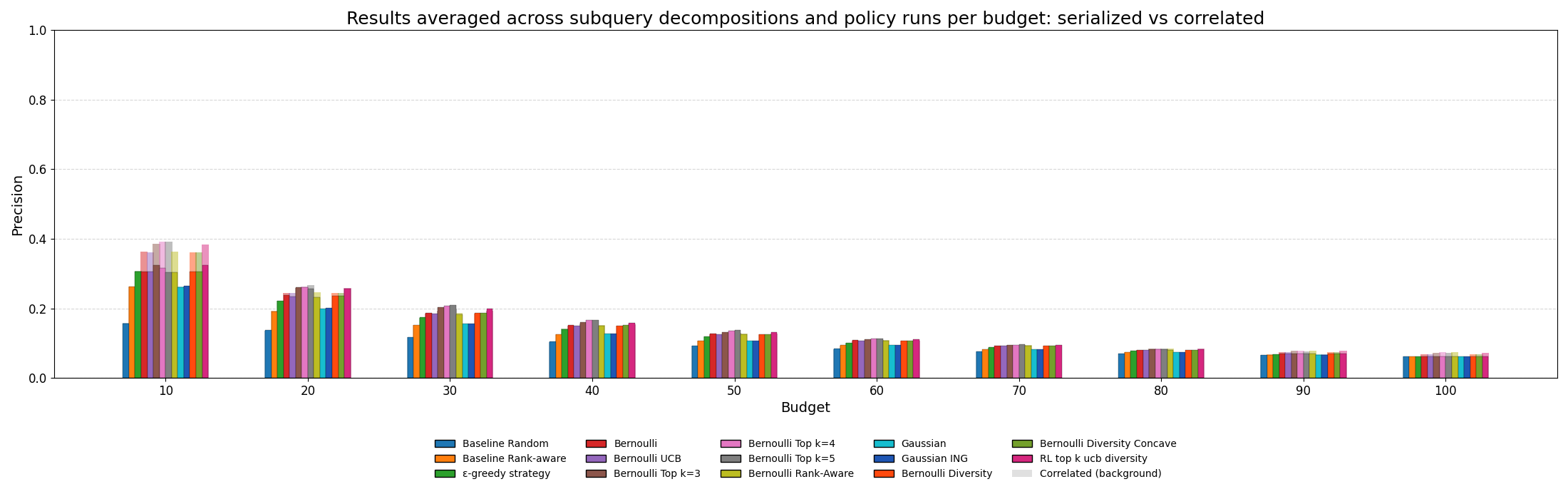}
    \caption{Precision applying baselines and reward policies for query selection for different budgets over observed documents on the ResearchyQuestions dataset: serial vs correlated.}
    \label{fig:researchy_correlated}
\end{figure*}

Figures~\ref{fig:neuclir_precision_all}–\ref{fig:researchy_recall_all} report precision and recall across budgets for all baselines and bandit policies we introduce in Table \ref{tab:reward_functions}. Figure ~\ref{fig:researchy_correlated} compared serial and hierarchy-aware document selection. 

\begin{figure*}[t]
    \centering
    \includegraphics[width=\linewidth]{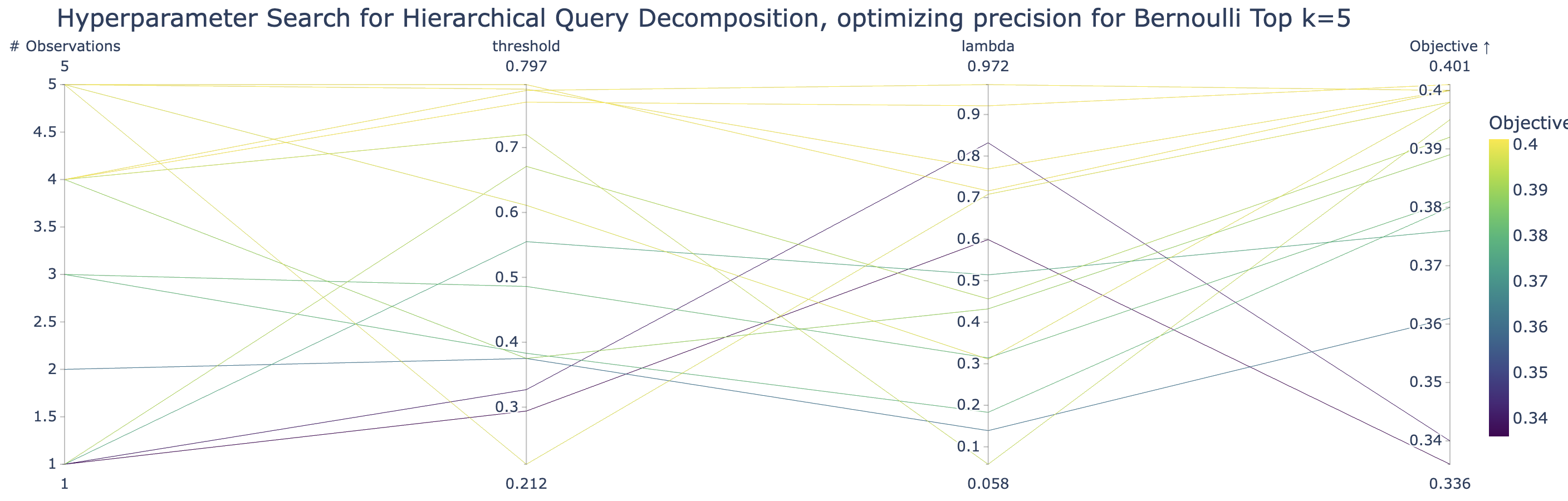}
    \caption{Visualization of hyperparameter search space with Bayesian optimization on the precision objective of the RL top k = 5 approach on the Researchy dataset. Each line corresponds to a trial with specific values of the three hyperparameters, colored according to the achieved objective.}
    \label{fig:bayesian_lookup}
\end{figure*}

\end{document}